\documentclass[runningheads]{llncs}

\usepackage{import}

\usepackage{amsmath}
\usepackage{amssymb}
\usepackage{amsfonts}

\usepackage{bm}
\usepackage{babel}
\usepackage[utf8]{inputenc}

\usepackage[%
  square,        %
  comma,         %
  numbers,       %
  sort&compress, %
]{natbib}

\usepackage{acronym}
\usepackage{booktabs}
\usepackage{multirow}
\usepackage{stackengine}
\usepackage{microtype}

\usepackage[ruled,vlined]{algorithm2e}
\usepackage{graphicx}
\usepackage{textcomp}
\usepackage[dvipsnames]{xcolor}

\usepackage{wrapfig}
\usepackage{subcaption}

\usepackage{iftex}
\usepackage{hyperref}
\hypersetup{%
  linkcolor  = MidnightBlue,
  citecolor  = MidnightBlue,
  urlcolor   = MidnightBlue,
  colorlinks = true,
}

\usepackage{xfrac}

\usepackage{mathtools}  %

\ifPDFTeX%
    \input glyphtounicode%
\else%
\fi%

\DeclareUnicodeCharacter{0229}{\,e}
\DeclareUnicodeCharacter{0327}{\,c}
\DeclareUnicodeCharacter{0303}{\~a}

\usepackage{tikz}

\newcommand\job[2]{%
  \begin{tikzpicture}%
    \draw[fill=#1, line width=.5pt] rectangle(#2, 1.3ex);
  \end{tikzpicture}%
}

\definecolor{JobRed}{HTML}{F8CECC}
\definecolor{JobGreen}{HTML}{D5E8D4}
\definecolor{JobOrange}{HTML}{FFE6CC}

\newcommand{\jobone}{\job{JobGreen}{2.2ex}}
\newcommand{\jobtwo}{\job{JobOrange}{3.3ex}}
\newcommand{\jobthree}{\job{JobRed}{4.4ex}}
\newcommand\emptycluster{
  \begin{tikzpicture}[scale=0.5, baseline=+0mm]
    \draw (0, 0) -- +(0mm, +4.2mm) ;
    \draw (0, 0) -- +(+8mm, 0mm) ;
  \end{tikzpicture}
}
\newcommand\onejobincluster{
  \begin{tikzpicture}[scale=0.5, baseline=+0mm]
    \draw[fill=JobGreen, line width=.3pt] rectangle(2.5mm, 2mm);
    \draw (0, 0) -- +(0mm, +4.5mm) ;
    \draw (0, 0) -- +(+8mm, 0mm) ;
  \end{tikzpicture}
}
\newcommand\twojobsincluster{
  \begin{tikzpicture}[scale=0.5, baseline=+0mm]
    \draw[fill=JobGreen, line width=.3pt] rectangle(2.5mm, 2mm);
    \draw[fill=JobOrange, line width=.3pt, yshift=2mm] rectangle(4.0mm, 2mm);
    \draw (0, 0) -- +(0mm, +4.5mm) ;
    \draw (0, 0) -- +(+8mm, 0mm) ;
  \end{tikzpicture}
}

\acrodef{AI}[AI]{Artificial Intelligence}
\acrodef{DRL}[DRL]{Deep Reinforcement Learning}
\acrodef{IL}[IL]{Imitation Learning}
\acrodef{LFD}[L\textsc{f}D]{Learning from Demonstration}
\acrodef{DQN}[DQN]{Deep Q Network}
\acrodef{HPC}[HPC]{High Performance Computing}
\acrodef{ML}[ML]{Machine Learning}
\acrodef{MLP}[MLP]{Multi-Layer Perceptron}
\acrodef{ReLU}[ReLU]{Rectified Linear Unit}
\acrodef{RL}[RL]{Reinforcement Learning}
\acrodef{MDP}[MDP]{Markov Decision Process}
\acrodefplural{MDP}[MDPs]{Markov Decision Processes}
\acrodef{SWF}[SWF]{Standard Workload Format}
\acrodef{TD}[TD]{Temporal-Difference}
\acrodef{ALE}[\textsc{ale}]{Arcade Learning Environment}
\acrodef{FCFS}[FCFS]{First-Come First-Serve}
\acrodef{FIFO}[FIFO]{First In First Out}
\acrodef{TDD}[\textsc{tdd}]{Test-Driven Development}

\acrodef{A2C}[A2C]{Advantage Actor-Critic}
\acrodef{PPO}[\textsc{ppo}]{Proximal Policy Optimization}
\acrodef{SAC}[SAC]{Soft Actor-Critic}
\acrodef{AC}[AC]{Actor-Critic}
\acrodef{CNN}[\textsc{cnn}]{Convolutional Neural Network}
\acrodef{DeepRM}[\textsc{DeepRM}]{Deep Resource Management}
\acrodef{DRaS}[\textsc{DRaS}]{Deep Reinforcement agent for Scheduling in HPC}

\begin{document}

\title{On the impact of MDP design for Reinforcement Learning agents in Resource Management\thanks{%
  \color{red}
  Author post-print. Accepted for publication at BRACIS 2021.
  When the final authenticated publication is made available online, this copy will be updated.
}}
\titlerunning{On the impact of MDP design for Resource Management RL agents}

\author{%
  Renato Luiz de Freitas Cunha\orcidID{0000-0002-3196-3008} \\\and%
  Luiz Chaimowicz\orcidID{0000-0001-8156-9941}%
}

\institute{%
    Programa de Pós Graduação em Ciência da Computação,\\
    Universidade Federal de Minas Gerais (PPGCC-UFMG),\\
    Belo Horizonte, MG, Brazil\\
    \email{\{renatoc,chaimo\}@dcc.ufmg.br}
}

\maketitle              %

\begin{abstract}
  The recent progress in Reinforcement Learning applications to Resource
  Management presents \acp{MDP} without a deeper analysis of the impacts of
  design decisions on agent performance. In this paper, we compare and contrast
  four different MDP variations, discussing their computational requirements
  and impacts on agent performance by means of an empirical analysis. We
  conclude by showing that, in our experiments, when using Multi-Layer
  Perceptrons as approximation function, a compact state representation allows
  transfer of agents between environments, and that transferred agents have
  good performance and outperform specialized agents in 80\% of the tested
  scenarios, even without retraining.

\end{abstract}

\vspace*{-1cm}
\section{Introduction}\label{sec:introduction}

\ac{DRL} has the potential of finding novel solutions to complex problems,
as outlined by recent progress in diverse areas such as Control of Gene
Regulatory Networks~\cite{nishida2018control}, adaptive video
acceleration~\cite{ramos2020straight}, and
management of computational resources~\cite{mao2016resource}.

Resource management, the process by which we map computational resources to the
tasks and jobs (programs) that require them, in particular, is an area in which recent learning
approaches have demonstrated superior performance over classical algorithms and
optimization techniques. Still, we see that, in recent work, each approach
defines their own \ac{MDP} formulations, with different design decisions. Thus, the
literature lacks an analysis of the impact of certain decisions on agent
performance.

In this paper, we investigate what happens to agent performance as we modify an
\ac{MDP}, observing the impacts when we change the state representation, the
transition function, and when we shape the reward signal, performing an
empirical investigation using open-source software from the deep learning and
\ac{RL} communities.

Our main contribution is the formulation and analysis of a set of \acp{MDP}
derived from one in the literature~\cite{mao2016resource} that allows agents to
learn faster, and to perform transfer learning with various function
approximation methods. We also show that doing so does not degrade performance
in the task.

The rest of this paper is organized as follows: In
Section~\ref{sec:related-work}, we describe the papers that influenced this
one, together with other deep \ac{RL} work for resource management. In
Section~\ref{sec:background}, we describe the theoretical background with an
ongoing example applied to a resource management problem. In
Section~\ref{sec:methodology}, we describe our methods and the proposed
extensions to a base \ac{MDP}. In
Section~\ref{sec:experiments}, we describe our experimental framework, along
with the experiments designed to evaluate our \acp{MDP}. In
Section~\ref{sec:conclusion} we present our concluding remarks and a brief
discussion of consequences of the work described here.
\section{Related Work}\label{sec:related-work}

In recent years, interest in Deep \acf{RL} applied to executing the scheduling
of computing jobs was probably inspired by the
\ac{DeepRM}~\cite{mao2016resource} agent and environment. \ac{DeepRM} presented
an approach of using Policy Gradients to schedule jobs based on CPU and memory
requirements. \ac{DeepRM}'s approach uses images to represent jobs, and
a window of jobs from which it can choose which job to schedule next. It was
shown that \ac{DeepRM} can learn to schedule based on different metrics.
\citet{domeniconi2019cush} proposed CuSH, a system that built on \ac{DeepRM} to
schedule for CPUs and GPUs, but proposed a hierarchical agent by introducing
an additional \ac{CNN} that chooses which job is going to be scheduled next,
and then uses a policy network to choose the scheduling policy to use with the
previously selected job. It is important to highlight a major difference
between \ac{DeepRM} and CuSH\@: whereas \ac{DeepRM} learns the scheduling
\emph{policy} itself, CuSH is essentially a classifier, which chooses between
two existing policies. This means that, even without training, CuSH's behavior
is more stable than that of \ac{DeepRM}, since \ac{DeepRM}-style schedulers
might get stuck in local minima, as reported by~\citet{cunha2020towards}, who
investigated the behavior of \ac{DeepRM}-style agents when trained with
state-of-the-art \ac{RL} algorithms such as \ac{A2C} and \ac{PPO}, and proposed
an OpenAI Gym environment for easier evaluation of \ac{RL} agents for job
scheduling.

Another agent that has been proposed recently, and that learns the scheduling
policy itself, is \textsc{RLscheduler}~\cite{zhang2020rlscheduler}.
\textsc{RLscheduler} is a \ac{PPO}-based agent with a fully convolutional
neural network for scoring jobs in a fixed window of size 128. The major
innovation in \textsc{RLscheduler} is in the training setting, in which the
authors combine synthetic workload traces with real workload traces to present
the agent with ever more difficult settings, similar to learning a curriculum
of tasks.

Similarly to CuSH, other agents that use a classification approach have been
proposed. A recent one is the \ac{DRaS}~\cite{fan2021deep}, which classifies
jobs in three categories: \emph{ready}, \emph{reserved}, and \emph{backfilled}.
After this classification step, the cluster scheduler takes the output of this
classification and allocates jobs accordingly (for example, by reserving slots
in the future for reserved jobs, scheduling immediately ready jobs, and finding
``holes'' in the schedule for backfilled jobs). \ac{DRaS} uses a five-layer
\ac{CNN} that works in two levels, with the first level selecting jobs for
immediate and reserved execution, and the second layer for backfilled
execution.

\medskip

None of the papers mentioned above discuss the impact of their design
decisions, resorting to only comparing their results with existing algorithms.
In this paper, we aim to analyze how decisions in \ac{MDP} design impact
\ac{DeepRM}-style algorithms, and how they impact both computational
performance and scheduling performance.
\section{Background}\label{sec:background}

In this section, we describe the background needed to understand the techniques
and methodology presented in this paper. To help in understanding,  we will use a resource management problem as running example throughout
this section.

\subsection{Batch job scheduling}

The primary goal of a job scheduler is to manage the job queue and coordinate
execution of jobs in \ac{HPC} clusters, while matching jobs to resources in an
efficient way. In a discrete time setting, at each time step, zero or more jobs
may arrive in the queue for processing, and the scheduler's job is to allocate
jobs to resources while satisfying their resource requirements. The job
scheduler guarantees jobs execute when requested resources are available, and
usually guarantee there won't be oversubscription of resources\footnote{%
    Some schedulers allow for oversubscription of memory resources in their
    default configuration, inspired by the fact that jobs don't use peak memory
    during their complete lifetimes.
}. Given this primary goal, secondary goals vary between schedulers and
\ac{HPC} facilities, depending on whether the hosting institution prefers to
satisfy the needs of individuals submitting jobs, or the whole group of
users~\cite{feitelson1996toward}.

When optimization of response time is a subgoal, it is usually modeled as the
minimization of the average response time, with response time used as a synonym
to turnaround time: the difference between the time a job was submitted to the
time it \emph{completed} execution. A metric commonly used to evaluate this is
the \emph{slowdown} of a job, which, for job $j$ is defined as
\begin{equation}
    \mathrm{slowdown}(j)=\frac{(t_f(j)-t_s(j))}{t_e(j)} =
    \frac{t_w(j)+t_e(j)}{t_e(j)} =
    \frac{1}{t_e(j)}\left(\sum_{i=1}^{t_w(j)}1+\sum_{i=1}^{t_e(j)}1\right)\text{,}
    \label{eq:slowdown}
\end{equation}
where $t_s(j)$ is the time job $j$ was submitted, $t_e(j)$ is the time it took
to execute job $j$, and $t_f(j)$ is the finish time of job $j$. The equality in
the middle holds because the wait time, $t_w$, of a job $j$ is defined as
$t_w(j)=t_f(j)-(t_e(j)+t_s(j))$.

\begin{figure}[htb]
    \centering
    \includegraphics[width=0.7\linewidth]{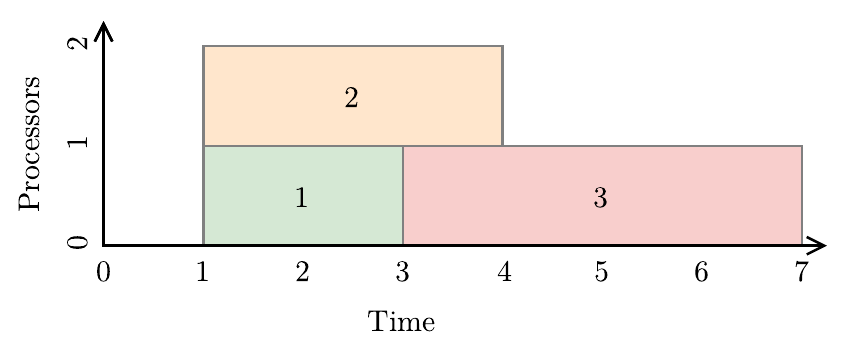}
    \caption{A possible schedule when three jobs arrive in a scheduling
    system at discrete time step 1 and no more jobs are submitted to the system
    at least until time step 7, the last one shown in the figure.}%
    \label{fig:schedule}
\end{figure}

Consider the case of three batch jobs, $j_1=$ \jobone, $j_2=$ \jobtwo, and
$j_3=$ \jobthree{}, submitted to a scheduling system with two processors, and
that the three jobs were submitted ``between'' time step 0 and 1, such that,
when transitioning from the first time step to the second, now there are three
jobs waiting. Also consider that, for these jobs, the generated schedule
is the one displayed in Figure~\ref{fig:schedule}. As shown in the figure, the
jobs execute for two, three and four time steps respectively, and all of them
use a single processor.

The reader should observe that different schedules can yield substantially
different values of (average) slowdown. For example, the schedule shown in
Figure~\ref{fig:schedule} has an average slowdown equal to
$\sfrac{1}{3}\sum_{i=1}^{3}\mathrm{slowdown}(j_i) = \sfrac{1}{3}(1
+ 1 + \sfrac{3}{2}) = \sfrac{7}{6}$, whereas, if we swapped $j_3$ with $j_1$,
and started $j_1$ soon after $j_2$ finished, the slowdown would be
$\sfrac{1}{3}(\frac{3+2}{2} + 1 + 1) = \sfrac{9}{6} = \sfrac{3}{2}$, a $\approx
29\%$ increase. Therefore, a scheduler should choose job sequences wisely,
otherwise its performance can be degraded.

In this paper, we will focus our discussion on what happens when an \ac{RL}
system tries to minimize the average slowdown, but our conclusions are general
and apply to other metrics and problems as well.

\subsection{Deep Reinforcement Learning and Job Scheduling}

In a \acf{RL} problem, an agent interacts with an unknown environment in which
it attempts to optimize a reward signal by sequentially observing the
environment's state and taking actions according to its perception.  For each
action, the agent receives a reward. Thus, in the end, we want to find the
sequence of actions that maximizes the total reward, as we will detail in the
next paragraphs.

\ac{RL} formalizes the problem as a \acf{MDP} represented by a tuple
$\mathcal{M} = \langle\mathcal{S}, \mathcal{A}, \mathcal{R},
\mathcal{T}, \rho, \gamma\rangle$\footnote{%
    Some authors leave the $\gamma$ component out of the definition of the
    \ac{MDP}. Leaving it in the definition yields a more general formulation,
    since it allows one to model continuous (non-ending) learning settings.
}.  At each discrete time step $t$ the agent is in
state $S_t \in \mathcal{S}$. From $S_t$, the agent takes an action $A_t \in
\mathcal{A}$, receives reward $R_{t+1} \in \mathcal{R}$ and ends up in state
$S_{t+1} \in \mathcal{S}$. Therefore, when we assume the first time step is
$0$, the interaction between agent and environment create a sequence $S_0,
A_0, R_1, S_1, A_1, R_2, \ldots$ of states, actions and rewards. To a specific
sequence $S_0, A_0, R_1, S_1, A_1, R_2, \ldots$ of states, actions, and
rewards we give the name of trajectory, and will denote such sequences by
$\tau$. The transition from state $S_t$ to $S_{t+1}$ follows the probability
distribution defined by $\mathcal{T} : \mathcal{S} \times \mathcal{A}
\rightarrow \mathcal{S}$ or, in an equivalent way, $\mathcal{T}$ gives the
probabiliy of reaching any new state $s'$ when taking action $a$ when in state
$s$: $p(s'|s, a) = p(S_{t+1}=s' | S_t=s, A_t=a)$. $\rho$ is a distribution of
initial states, and $\gamma$ is a parameter $0 \leq \gamma \leq 1$, called the
discount rate.  The discount rate models the present value of future rewards.
For example, a reward received $k$ steps in the future is worth only
$\gamma^k$ now. This discount factor is added due to the uncertainty in
receiving rewards and is useful for modeling stochastic environments. In such
cases, there is no guarantee an anticipated reward will actually be received
and the discount rate models this uncertainty.

To map our presentation of \ac{RL} into our problem of job scheduling, we
consider $\rho(\emptycluster) = 1$ (the only possible initial state is the
empty cluster), with the first state consisting of the empty cluster, with
no jobs in the system, $S_0=\langle\emptycluster\rangle$ and
$A_0 = \emptyset$, since there is no job to schedule.

Recall our discussion
about classifying jobs to be processed by different policies, \emph{versus}
choosing the next job to enter the system. In this paper, we are modeling
an \ac{MDP} in which the next job is chosen by the agent, so the agent is learning
a scheduling policy.
In our example, one can obtain a reward function by using the sequential version
of slowdown, shown in the rightmost equality of~\eqref{eq:slowdown}, such that
the reward at each time step is given by the sum of the current slowdown for
all jobs in the system: $\mathcal{R}=-\sum_{j\in\mathcal{J}}\sfrac{1}{t_e(j)}$.
When the reward function is such that it computes the online version of
slowdown for \emph{all jobs in the system}, if $A_1=\emptyset$,
$R_2=\sfrac{1}{2}+\sfrac{1}{3}+\sfrac{1}{4}$. Moreover, if jobs $j_1$, $j_2$,
and $j_3$ are chosen in sequence, the next state, shown in
Fig.~\ref{fig:schedule}, will be given by sequentially applying the transition
function $\mathcal{T}$ as $\mathcal{T}(\emptycluster,
\jobone)\mathcal{T}(\onejobincluster, \jobtwo)\mathcal{T}(\twojobsincluster,
\jobthree)$. If the episode finished immediately after the state shown in
Fig.~\ref{fig:schedule}, the trajectory $\tau_1$ would be given by $\tau_1
= \langle S_0=\emptycluster, A_0=\jobone,  R_1=0,
S_1=\onejobincluster, A_1=\jobtwo, R_2=0, S_2=\twojobsincluster,
\ldots \rangle$\footnote{%
    The value shown for $R_2$ might contradict the previous discussion, but the
    \ac{MDP} is set in a way that, \emph{when jobs are scheduled successfully},
    $R_{t+1}=0$.
}.

The reward signal encodes all of the agent's goals and purposes, and the
agent's sole objective is to find a policy $\pi_\theta$ parameterized by
$\theta$ that maximizes the expected return
\begin{equation}
    G(\tau) = R_{1} + \gamma R_{2} + \cdots + \gamma^{T-1} R_{T} =
    \sum_{t=0}^{T-1}\gamma^{t}R(S_t, A_t\sim\pi_\theta(S_t))\text{,}
    \label{eq:return}
\end{equation}
which is the sum of discounted rewards encountered by the agent. When $T$ is
unbounded, $\gamma < 1$. Otherwise, Equation~\eqref{eq:return} would diverge.
In our example, a deterministic policy that always chose the smallest job first
would yield $\pi_\theta(\langle\emptycluster, \jobone,
\jobtwo, \jobthree\rangle)=\jobone$, while a stochastic policy would assign
a probability to each job, and either choose the one with highest probability
or sample from the jobs according to that distribution. In practice, when
neural networks are used for approximation, the last layer of the neural
network is usually a softmax, so that each action gets a number that can be
interpreted as a probability.  As mentioned before, in the example in this
section, rewards are based on the negative online slowdown and, therefore,
returns will also depend on the slowdown. The policy $\pi$ is a mapping from
states and actions to a probability of taking an action $A$ when in state $S$,
and the parameters $\theta$ relate to the approximation method used by the
policy\footnote{%
    In our example, for each job $j_i$, in time step $1$, $\pi$ would
    give the probabilities of choosing each job given an empty cluster:
    $\pi(\jobone|\emptycluster)$, $\pi(\jobtwo|\emptycluster)$, and
    $\pi(\jobthree|\emptycluster)$ such that, by total probability,
    $\pi(\jobone|\emptycluster) + \pi(\jobtwo|\emptycluster) +
    \pi(\jobthree|\emptycluster) = 1$.
}. Popular function approximators include linear combinations of
features~\cite{liang2016state} and neural
networks~\cite{tesauro1994td,silver2018general}.

\subsection{Policy gradients}\label{sec:background-pg}

In this section we present the main optimization method we use to find
policies: policy gradients. As implied by the name, we compute
gradients of policy approximations, and use them to find
better parameters for those functions.

Formally, we generalize policies to define distributions over trajectories with
\begin{equation}
    \phi_\theta(\tau) = \rho(S_0)\prod_t\pi_\theta(A_t|S_t)
    \underbrace{\mathcal{T}(S_{t+1}|S_t, A_t)}_{\text{Environment}}\text{,}
    \label{eq:trajectories}
\end{equation}
in which $\pi_\theta$ is being optimized by the agent, and $\rho$ and $\mathcal{T}$
are provided by the environment. What~\eqref{eq:trajectories} says is that we can
assign probabilities to any trajectory, since we know the distribution of initial
states $\rho$, and we know that the policy will assign probabilities to actions
given states, and that, when such actions are taken, the environment will sample
a new state for the agent.
When we do so, we can define an optimization objective to find the optimal set
of parameters
\begin{equation}
    \theta^* = \arg \max_\theta J(\theta) = \arg \max_\theta\int_\tau
    G(\tau)\phi_\theta(\tau)d\tau\text{,}
\end{equation}
where $J(\theta)$ is the performance measure given by the expected return of
a trajectory, which can be approximated by a Monte Carlo estimate
$\widehat{J(\theta)}=\sfrac{1}{N}\sum_{i}G(\tau_i)\phi_\theta(\tau_i)$\footnote{%
    Normalization is needed to approximate the average value of $\widehat{J(\theta)}$.
    Otherwise, $\widehat{J(\theta)}\rightarrow\infty$ as $N\rightarrow\infty$.
}. If we
construct $\widehat{J(\theta)}$ such that it is differentiable, we can
approximate $\theta^*$ by gradient ascent in $\theta$, such that
$\theta_{j+1}\leftarrow\theta_j+\alpha\nabla\widehat{J(\theta)}$, with $\alpha > 0$,
yielding
\begin{equation}
    \nabla_\theta J(\theta) \approx \frac{1}{N}\sum_i
    G(\tau_i)\nabla_\theta\log\phi_\theta(\tau_i);\tau_i\sim\pi_\theta\text{.}
    \label{eq:pg}
\end{equation}

By expanding~\eqref{eq:return} by one time-step, we get the update
$G(\tau)=R_{1} + \gamma G(\tau_1)$ or, more generally, $G(\tau_t)=R_{t+1}
+ \gamma G(\tau_{t+1})$, where $\tau_i$ indicates the trajectory $\tau$
starting from offset $i$. This update is usually written as $G_t=R_{t+1}+\gamma
v_\pi(S_{t+1})$, where $G_t$ is shorthand notation for $G(\tau_t)$, and
$v_\pi(S_{t+1})$ is the return when starting at state $S_{t+1}$ and following
policy $\pi$ (which generated trajectory $\tau$). Another function related to
$v_\pi(S_t)$ is $q_\pi(S_t, A_t)$, which gives the return when starting at
state $S_t$ \emph{and} taking action $A_t$ then following policy $\pi$. With
these two functions, we can define a third one, which gives the relative
\emph{advantage} of taking action $A_t$ when in state $S_t$, defined as
$a_\pi(S_t, A_t) = q_\pi(S_t, A_t)-v_\pi(S_t)$, and called the advantage
function. As with the policy, $q_\pi$, $v_\pi$, and $a_\pi$ can also be
approximated and, thus, learned. When such an approximation is used, the
update~\eqref{eq:pg} becomes
\begin{equation}
    \nabla_\theta J(\theta) \approx \frac{1}{N}\sum_i
    \nabla_\theta\log\pi_\theta(A_t|S_t)\hat{a}_\pi(S_t, A_t); S_t, A_t\sim\pi_\theta\text{,}
    \label{eq:pg-advantage}
\end{equation}
where $\hat{a}_\pi$ is an approximation of $a_\pi$, and which can be further
split into two estimators as $\hat{a}_\pi=\hat{q}_\pi-\hat{v}_\pi$, with the
arguments $S_t$ and $A_t$ dropped for better readability. In this setting, the
$\pi_\theta$ approximator is called an actor, and the $\hat{a}_\pi$ approximator
is called a critic.

In the literature, we find techniques that
regularize updates~\cite{mnih2016asynchronous,schulman2017proximal}, but as
presented, equation~\eqref{eq:pg-advantage} is sufficient for understanding of
the techniques discussed in this paper.

\section{Methodology}\label{sec:methodology}

Although the discussion in the previous section is helpful for conceptualizing
the problem we are interested in, it is not enough to help us \emph{implement}
a solution, since it does not specify when the agent is invoked for learning,
how a state is actually represented, nor how rewards are computed for each
action. In this section, we will detail our design decisions, and will
elaborate on what changes are required to assess the impact of said decisions
in \ac{RL} performance. We begin by describing the base \ac{MDP}, and then we
will describe incremental changes that can be made to the environment so that
it may become faster to compute, and easier to learn, leading to faster
convergence.

We implemented the base \ac{MDP} and each incremental change discussed in this
section. Then, we evaluated all implementations, observing both convergence
performance and final agent performance in the task of scheduling \ac{HPC}
jobs.

\subsection{The base, image-like MDP}\label{sec:methodology-dense}

\begin{wrapfigure}[18]{r}{.5\textwidth}
  \begin{center}
    \def\svgwidth{.5\textwidth}
    \vspace*{-3.5em}
\begingroup%
  \makeatletter%
  \providecommand\color[2][]{%
    \errmessage{(Inkscape) Color is used for the text in Inkscape, but the package 'color.sty' is not loaded}%
    \renewcommand\color[2][]{}%
  }%
  \providecommand\transparent[1]{%
    \errmessage{(Inkscape) Transparency is used (non-zero) for the text in Inkscape, but the package 'transparent.sty' is not loaded}%
    \renewcommand\transparent[1]{}%
  }%
  \providecommand\rotatebox[2]{#2}%
  \newcommand*\fsize{\dimexpr\f@size pt\relax}%
  \newcommand*\lineheight[1]{\fontsize{\fsize}{#1\fsize}\selectfont}%
  \ifx\svgwidth\undefined%
    \setlength{\unitlength}{134.27887344bp}%
    \ifx\svgscale\undefined%
      \relax%
    \else%
      \setlength{\unitlength}{\unitlength * \real{\svgscale}}%
    \fi%
  \else%
    \setlength{\unitlength}{\svgwidth}%
  \fi%
  \global\let\svgwidth\undefined%
  \global\let\svgscale\undefined%
  \makeatother%
  \begin{picture}(1,0.50465229)%
    \lineheight{1}%
    \setlength\tabcolsep{0pt}%
    \put(0,0){\includegraphics[width=\unitlength,page=1]{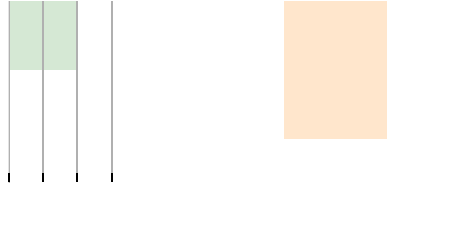}}%
    \put(0.49721107,0.03363329){\color[rgb]{0,0,0}\makebox(0,0)[lt]{\lineheight{1.25}\smash{\begin{tabular}[t]{l}Slots\end{tabular}}}}%
    \put(0,0){\includegraphics[width=\unitlength,page=2]{img/state.pdf}}%
    \put(0.41031137,0.07357579){\color[rgb]{0,0,0}\makebox(0,0)[lt]{\lineheight{1.25}\smash{\begin{tabular}[t]{l}1\end{tabular}}}}%
    \put(0.70587764,0.07563713){\color[rgb]{0,0,0}\makebox(0,0)[lt]{\lineheight{1.25}\smash{\begin{tabular}[t]{l}2\end{tabular}}}}%
    \put(0.1297461,0.06675814){\color[rgb]{0,0,0}\makebox(0,0)[t]{\lineheight{1.25}\smash{\begin{tabular}[t]{c}Cluster\\Processors\end{tabular}}}}%
    \put(0.80182648,0.06734041){\color[rgb]{0,0,0}\makebox(0,0)[lt]{\lineheight{1.25}\smash{\begin{tabular}[t]{c}Backlog\\Slots\end{tabular}}}}%
  \end{picture}%
\endgroup%
   \end{center}
  \vspace*{-5mm}\caption{%
    Dense state representation with images representing state.
    In the figure, there is one job in execution (with two processors for
    the next two time steps), three waiting jobs in total, two of them within
    window $W=2$, one using one CPU for at least five time steps, and another
    using three CPUs for four time steps. Details for the third job, in the
    backlog are omitted.
  }\label{fig:dense-state}
\end{wrapfigure}

We begin by following the design of DeepRM~\cite{mao2016resource},
summarized here, and exemplified in Fig.~\ref{fig:dense-state}. We start by
representing states as images whose height corresponds to a look to
a time horizon of $H$ time-steps ``into'' the future, and the width comprising:
the number of processors in the system and their occupancy state, a window of
configurable size $W$ (in the Figure, $W=2$) containing the first $W$ jobs in
the wait queue times the number of processors in the system, and a column
vector indicating jobs in a ``backlog''\footnote{%
  Jobs in the wait queue that the agent \emph{cannot} choose to schedule.
}.
If there are more jobs in the system that can fit the window and the
backlog, they are omitted from the state representation\footnote{%
  Truncating the list of jobs violates the Markov property, since once it
  overflows, the agent cannot know how many jobs are in the system.
}.

For the action, the agent can either choose to schedule a job from one of the $W$
slots, or it can refuse to schedule a job, totalling three possible actions in
the example of Fig.~\ref{fig:dense-state}. Regarding \emph{when} actions are taken,
the \ac{MDP} was built in such a way that agents see every simulation time step
and ``intermediate'' time-steps as well: when a job is scheduled, there is
a state change in the \ac{MDP}, with the job moving to the in-use processors,
and the queue being re-organized so that all slots in window $W$ are filled.

In the base \ac{MDP}, whenever a job is scheduled, the agent receives a reward
of zero. In all other cases, the reward is given by the negative online
slowdown~\eqref{eq:slowdown}: $\mathcal{R}=-\sum_{j\in\mathcal{J}}\frac{1}{t_e(j)}$.
For a detailed description of the environment, we direct the reader
to~\citet{mao2016resource}, who first introduced it.

\subsection{Compact state representation}\label{sec:methodology-compact}

The first realization we had was that the state representation in the base \ac{MDP}
is wasteful, in the sense that one can reduce the size of the state without
losing information. Particularly when working with larger clusters, or with
a larger number of job slots, it may be the case that full trajectories take
too much space in memory, reducing the computational per\-for\-mance of
learning agents. Due to that, and based on a set of features found in the
literature for machine learning with \ac{HPC} jobs~\cite{cunha2017job}, we
devised a set of features that can represent states in a compact way. In our
new state representation, jobs in the queue are represented by the features
shown in Table~\ref{tab:tabular-features}, where ``work'' is computed by
multiplying the number of processors a job requires by the time it is expected
to run, and cluster features are a pair that indicates the number of processors
in use, and the number of free processors. The features related to the cluster
state still use a time horizon $H$ but instead of using a matrix, we used
a pair of integers representing how many processors are in use, and how many
processors are free in a given time-step. As an example, assuming the job in
the cluster was submitted at time 1, the job in slot 1 was submitted at time 2,
and the job in slot 2 was submitted in time 3, the state shown in
Fig.~\ref{fig:dense-state} can be fully described by the concatenation of
vectors with cluster state $\langle (2, 1), (2, 1), (0,
3), (0, 3), (0, 3) \rangle$, jobs in window $W$ $\langle (1, 5, 1, 0, 0, 1, 6,
0), (2, 4, 3, 1, 5, 1, 4, 0) \rangle$ and backlog $\langle 1 \rangle$\footnote{
  Parentheses group elements. In the first vector, there are five parenthesized
  pairs to indicate the time horizon of 5, and two parenthesized elements to
  represent job slows in window $W$.
}. The features in the jobs slots are presented in the same order as the ones
shown in Table~\ref{tab:tabular-features}.

\begin{table}[htpb]
  \vspace*{-5mm}
    \centering
    \caption{%
      Job features in a compact state representation.
    }\label{tab:tabular-features}
    \scriptsize
    \begin{tabular}{ll}
        \toprule
        Feature              & Description \\
        \midrule
        Submission time      & Time at which the job was submitted \\
        Requested time       & Amount of time requested to execute the job \\
        Requested processors & Number of processors requested at the submission time \\
        Queue size           & Number of jobs in the wait queue at job submission time \\
        Queued work          & Amount of work that was in the queue at job submission time \\
        Free processors      & Amount of free processors when the job was submitted \\
        Remaining work       & Amount of work remaining to be executed at job submission time \\
        Backlog              & The number of jobs waiting outside window $W$ \\
        \bottomrule
    \end{tabular}
\end{table}

A side-effect of using this new compact state representation is that, when $H$
and $W$ are fixed between different cluster configurations, learned features
are directly transferable between clusters even when using %
function approximation methods that depend on a fixed number of features.

\subsection{Sparse State Transitions}\label{sec:methodology-sparse}

Another deficiency we've identified in the base \ac{MDP} is that the agent sees
\emph{all} time-steps in the simulation, but this causes the agent to have to
take an action even when there is no good action to take. Consider, for
example, the case in which all resources are in use (there are no free
resources). In cases such as this, any action the agent takes will lead to the
same outcome: increasing the simulation clock, receiving negative rewards
related to the slowdown of the jobs, and having \emph{no} new jobs scheduled.
This will be repeated for all time steps between the start of the last job that
exhausted resources until the finish of the first job that frees them, causing
non-negative rewards to be more sparse, making the reinforcement signal noisier
and, therefore, harder to learn. The opposite is also true: if there are no
jobs waiting to be scheduled, no matter what the agent chooses, the outcome
will be the same: no jobs will be scheduled.

Due to that, we updated the environment to only call the agent and, therefore,
to only add states, actions and rewards to a trajectory, when it was possible
for the agent to take an action that could result in a job being scheduled. In
short, we change the transition function $\mathcal{T}(S_{t+1}\mid S_t, A_t)$ so
that all state transitions from $S_t$ to $S_{t+1}$ will always have at least one
job that may be scheduled by the agent in state $S_{t+1}$. We did not change the
initial state, though, so $\rho=\{\emptycluster\}$ still holds. This essentially
turns the \ac{MDP} into a semi-\ac{MDP}~\cite{sutton1999between}. To make our
formulation compatible with a semi-\ac{MDP}, we extend the reward function to
return zero in all intermediate states after successfully scheduling a job.

\subsection{Reducing the noise of the reward signal}\label{sec:methodology-noise}

Based on the idea of only showing the agent what it can use to learn and act,
we noticed that the reward signal could be further improved by, instead of
computing the online slowdown of all jobs in the system $\mathcal{J}$,
considering only the jobs that are in the waiting queue, and within the job
slots window $W$: the jobs that can be directly influenced by the agent's
actions. Therefore, we defined the set $\mathcal{W}$ that contains the subset
of jobs from $\mathcal{J}$ that are within the window $W$, and the reward
function became $\mathcal{R} = -\sum_{j\in\mathcal{W}}\frac{1}{t_e(j_i)}$ when
the action taken doesn't schedule a job, and 0 otherwise.

We evaluate the impact of the various \acp{MDP} on agent performance by
performing two sets of experiments, one in which we observe the impact of the
changes proposed in Sections~\ref{sec:methodology-compact}
through~\ref{sec:methodology-noise} (called \emph{Compact}, \emph{Sparse}, and
\emph{Reduced} respectively in the experiments) as opposed to the dense \ac{MDP}, and
another in which we observe the impact of using an event-based simulation and
bounded rewards both in dense and compact \acp{MDP}.
\section{Experiments}\label{sec:experiments}

In order to evaluate our methodology, we used open-source libraries to
implement both our agents and environment, with
\texttt{stable-base\-lines3}~\cite{stable-baselines3} providing the \ac{PPO}
agent and its training loop, and \texttt{sched-rl-gym}~\cite{cunha2020towards}
providing the simulator and environment implementation.

\begin{wraptable}[14]{r}{.30\linewidth}
  \footnotesize
  \vspace*{-11.0mm}
  \caption{%
    List of hyper-parameters used when training agents.
  }\label{tab:hyperparameters}
  \begin{tabular}{lr}
    \toprule
      Hyper-parameter & Value \\
    \midrule
      Learning rate & $10^{-4}$\\
      $n$ steps & $50$ \\
      batch size & $64$ \\
      Entropy coefficient & $10^{-2}$\\
      \textsc{gae} $\lambda$ & $0.95$\\
      Clipping $\epsilon$ & $0.2$\\
      Surrogate epochs & $10$\\
      $\gamma$ & $0.99$\\
      Value coefficient & $0.5$\\
    \bottomrule
  \end{tabular}
\end{wraptable}

All our experiments consisted of training a \ac{PPO} agent in the different
formulations of the previous section. We also fixed the neural network
architecture used for function approximation, consisting of a two-layer neural
network with 64 units in each layer, and with parameter sharing between policy
and value networks. The fixed number of units implies the image-like
representation will use more parameters, as it contains more data than the
compact representation. The hyper-parameters used for training the agent are
summarized in Table~\ref{tab:hyperparameters}. We performed no hyper-parameter
optimization, and used values found in the literature when training the
image-like agent. For a full description of \ac{PPO}, we direct the reader to
Schulman et al.~\cite{schulman2017proximal}.

We also maintained the environment specification fixed for all agent
evaluations and used $W=10$ job slots, with simulations of length $T=100$
time-steps and time horizon $H\in\{20, 60\}$. These two horizon values enable
us to contrast cases in which agents can see when jobs will complete, or not.
Regarding the workload, we used a workload generator from the
literature~\cite{mao2016resource,cunha2020towards}, which submitted a new job
with $30\%$ chance on each time step. Of these, a job had $80\%$ chance of
being a ``small'' job, and ``large'' otherwise. The number of processors $n_p$
was chosen in the set \{10, 32, 64\}, while the maximum job length (duration)
$d$ varied from \{15, 33, 48\} and the size of the largest job (number of
processors) $j_s$ came from the set \{10, 32, 64\}. In the workload generator, the
length of small jobs was sampled uniformly from $[1, \sfrac{d}{5}]$, and the
length of large jobs was sampled uniformly from $[\sfrac{2d}{3}, d]$. The
number of processors used by any job was sampled from $[\sfrac{n_p}{2}, n_p]$.

All agents were trained for three million time-steps as perceived by the agent.
This means that all agents will see the same number of states, and will take
the same number of actions, but the number of time steps in the underlying
simulation will vary, due to the event-based case becoming a semi-\ac{MDP}.
We evaluated agents with a thousand independent trials, reporting average values.

\begin{figure}[t]
  \centering
  \subcaptionbox{%
    $n_p=10, d=15, j_s=10$\label{fig:learning-curves-all-a}
  }{%
  \includegraphics[height=1in]{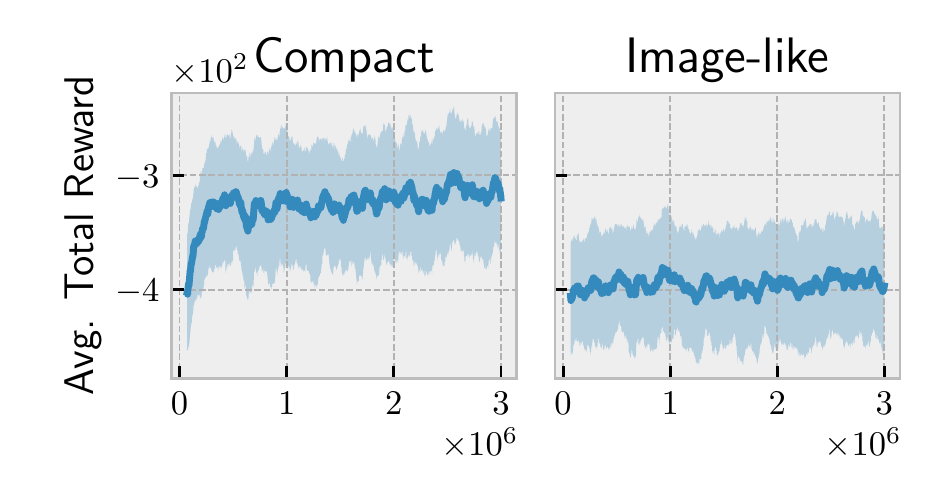}
  }
  \subcaptionbox{%
    $n_p=38, d=15, j_s=32$\label{fig:learning-curves-all-b}
  }{%
    \includegraphics[height=1in]{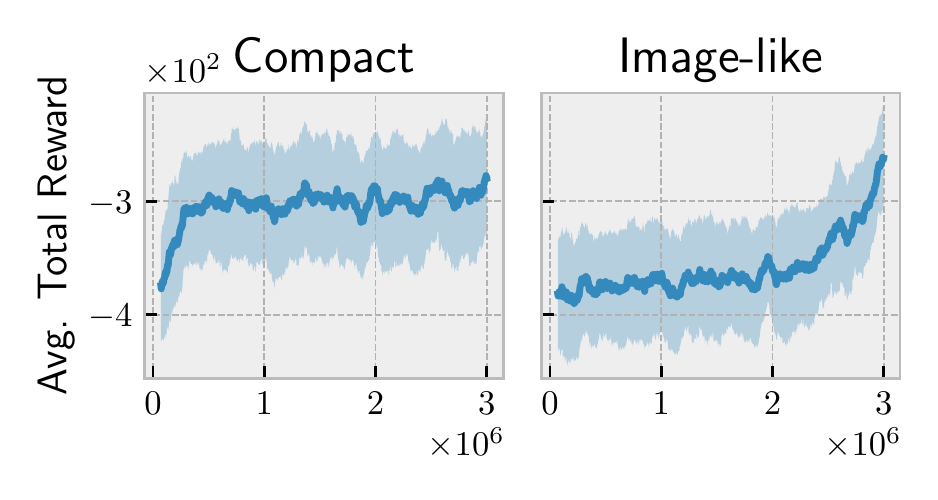}
  }
  \subcaptionbox{%
    $n_p=38, d=48, j_s=10$\label{fig:learning-curves-all-c}
  }{%
    \includegraphics[height=1in]{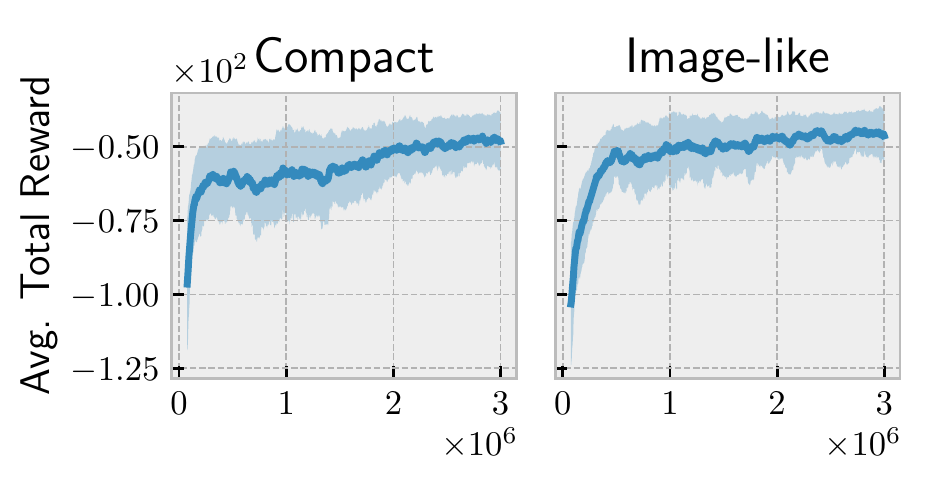}
  }
  \subcaptionbox{%
    $n_p=64, d=33, j_s=32$\label{fig:learning-curves-all-d}
  }{%
    \includegraphics[height=1in]{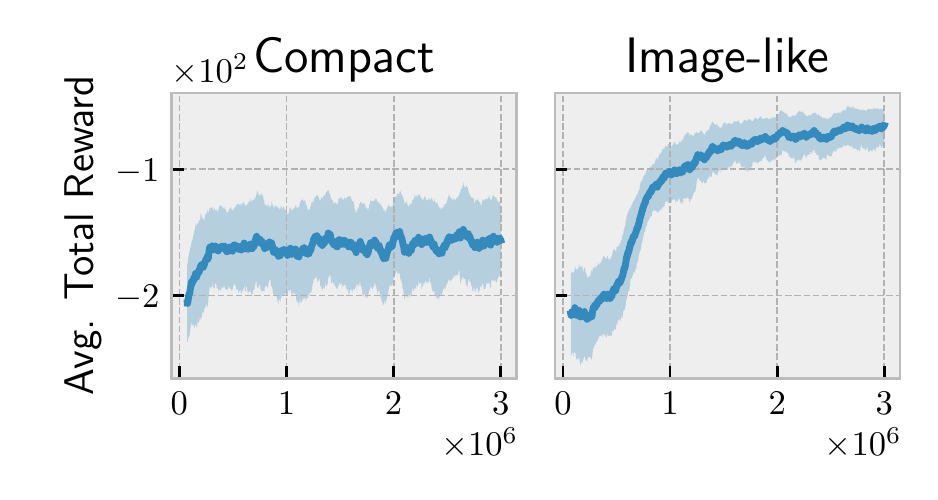}
  }
  \caption{%
    Learning curves for various scenarios with $H=20$ contrasting learning
    using a compact representation with learning with an image-like
    representation.  Curves are an average of six agents, with shaded areas
    representing one standard deviation, and show a moving average of total
    rewards received by the agents during training.
  }\label{fig:learning-curves-all}
\end{figure}

In Fig.~\ref{fig:learning-curves-all} we show a sampling of learning curves
comparing the learning performance of agents that were trained using the
image-like representation and the compact representation with rewards computed
from all jobs. The compact representation converges faster
than the image-like representation, probably due to its smaller number of
parameters. We also notice that although convergence is faster, the compact
representation is not necessarily better
(Fig.~\ref{fig:learning-curves-all-c},~\ref{fig:learning-curves-all-d}). There
doesn't seem to be a general rule, but we noticed that when jobs are shorter
(the $d$ parameter is smaller), the compact representation dominates
(Fig.~\ref{fig:learning-curves-all-a},~\ref{fig:learning-curves-all-b}). When
$d$ increases and most jobs use few processors ($j_s \ll n_p$), the compact
representation tends to have comparable performance with the image-like
representation (Fig.~\ref{fig:learning-curves-all-c}), whereas when jobs use
many processors \emph{and} have a longer duration, agents using the image-like
representation learn the environment better
(Fig.~\ref{fig:learning-curves-all-d}). For this set of experiments, the size
of the time horizon ($H$) doesn't impact the learning performance, as curves
obtained with $H=60$ (not shown) are indistinguishable from visual inspection
to the ones obtained with $H=20$. When evaluating agents, we performed t-tests
to check whether there was a difference in agent performance when using these
different $H$ values.  In other words, the null hypothesis was that performance
was equal, and the alternative hypothesis was that agent performance varied. In
this setting, the null hypothesis was rejected only $36.6\%$ of the time when
considering p-values~$\le 1\%$.

\begin{wraptable}[18]{r}{.47\linewidth}
  \centering
  \footnotesize
  \vspace*{-8.75mm}
  \caption{%
    Key to the scenarios presented in Fig.~\ref{fig:average-slowdown-1}.
    \emph{Procs.} refers to the number of processors in the cluster, \emph{Max
    Length} refers to the maximum job length, and \emph{Max Size} refers to the
    maximum number of processors used by jobs.
  }\label{tab:scenarios}
  \begin{tabular}{cccc}
    \toprule
      Scenario & Procs. & Max Length & Max Size \\
      \midrule
      1  & 10 & 15 & 10 \\
      2  & 10 & 48 & 10 \\
      3  & 38 & 15 & 32 \\
      4  & 38 & 33 & 32 \\
      5  & 38 & 48 & 32 \\
      6  & 64 & 15 & 64 \\
      7  & 64 & 33 & 32 \\
      8  & 64 & 33 & 64 \\
      9  & 64 & 48 & 32 \\
      10 & 64 & 48 & 64 \\
    \bottomrule
  \end{tabular}
\end{wraptable}

\begin{figure}[t]
  \centering
  \includegraphics[width=.6\linewidth]{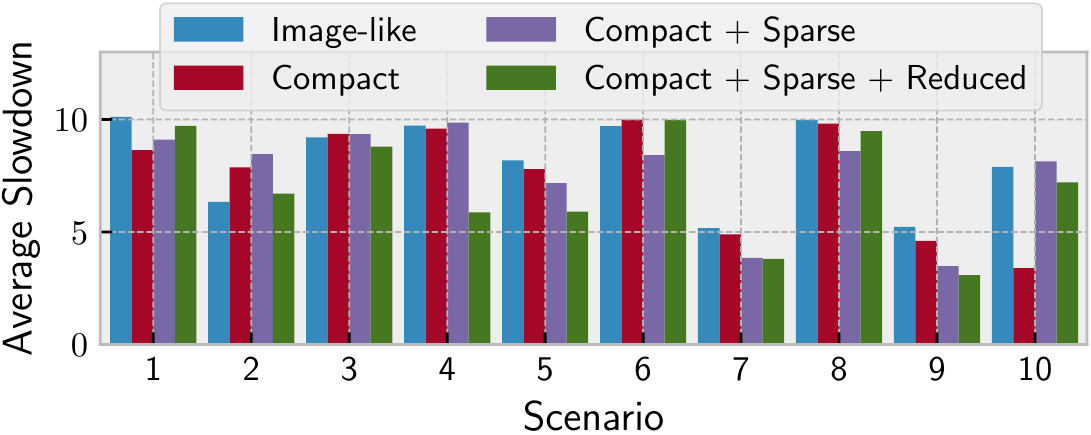}
  \caption{%
    Average slowdown for the various scenarios considered. Each bar represents
    a different instantiation of the (semi-)\acp{MDP}.  Average slowdowns were
    computed by averaging the slowdown of a thousand independent trials for
    each agent in each scenario.  All agents were evaluated with same workload
    and random seed. In the legend, \emph{image-like} corresponds to the base
    \ac{MDP}, \emph{compact} to the compact representation, \emph{sparse} to
    the sparse state transitions, and \emph{reduced} to the reduced set of jobs
    to compute rewards.
  }\label{fig:average-slowdown-1}
\end{figure}

When evaluating agents after one million iterations,
scheduling performance was similar between agents when the maximum
number of processors used by jobs was smaller (which implies
less parallelism).  Given job submission rates in all environments was the
same, clusters were less busy in these situations: as long as jobs
are scheduled, there shouldn't be significant differences in average slowdown,
due to smaller queues.

In Fig.~\ref{fig:average-slowdown-1}, with key to scenarions shown in
Table~\ref{tab:scenarios}, we show average slowdown of the agents for the
scenarios in which there was some variability in performance between agents.
From the figure, we see that, apart from scenarios 2 and 6, agent performance
in the ``Compact + Sparse + Reduced'' \ac{MDP} is not worse than that of the
image-like \ac{MDP}. Of these two, only the difference for scenario 6 is
statistically significant, with p-value $\le 5\%$ when performing a t-test
with alternative hypothesis of different distributions. For the cases where
``Compact + Sparse + Reduced'' agents are better, the results are statistically
significant (p-value $\le 5\%$) in scenarios 3, 4, 5, 7, and 9. Scenarios 2 and
6 are interesting, since they were configured to have shorter jobs of at most
15 time-steps, with scenario 1 having 10 processors, and scenario
6 having 64, both with jobs with the potential of using all cluster resources.

\begin{wrapfigure}[17]{r}{.48\textwidth}
  \vspace*{-6mm}
  \centering
  \includegraphics[width=\linewidth]{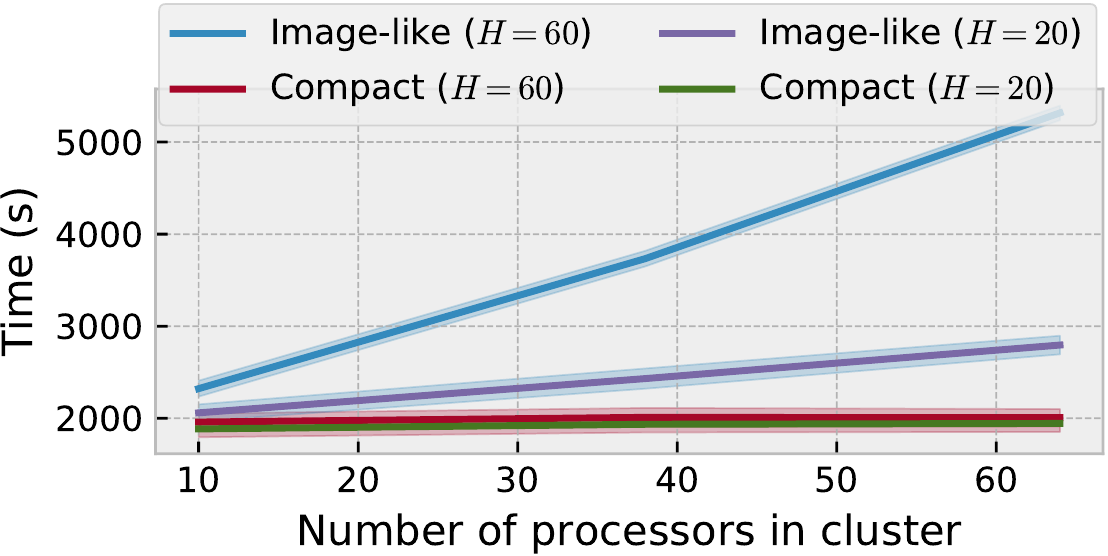}
  \caption{%
    Time needed to train agents for three
    million iterations. The shaded area represents one standard deviation.
    Increasing the time horizon increases the training time of compact agents by
    a constant factor, while it adds a linear factor to the training times of
    agents that use an image-like representation.
  }\label{fig:training-times}
\end{wrapfigure}

In Fig.~\ref{fig:training-times} we contrast the training times for the various
agents. As can be seen, training times for agents based on the dense \ac{MDP}
are highly variable, due to the fact that different \ac{MDP} configurations
result in different sizes of state representations, which impacts
training performance.
As an example, the image-like agent requires 301068,
1089548, and 1821708 parameters for the scenarios with 10, 38, and 64
processors, while all compact agents require a fixed number of parameters:
24332. Times were measured in a Linux 5.10.42 desktop with an NVIDIA GTX 1070
GPU and an i7--8700K processor using the \emph{performance} CPU
frequency-scaling governor.

\begin{figure}[htpb]
  \centering
  \includegraphics[width=0.8\linewidth]{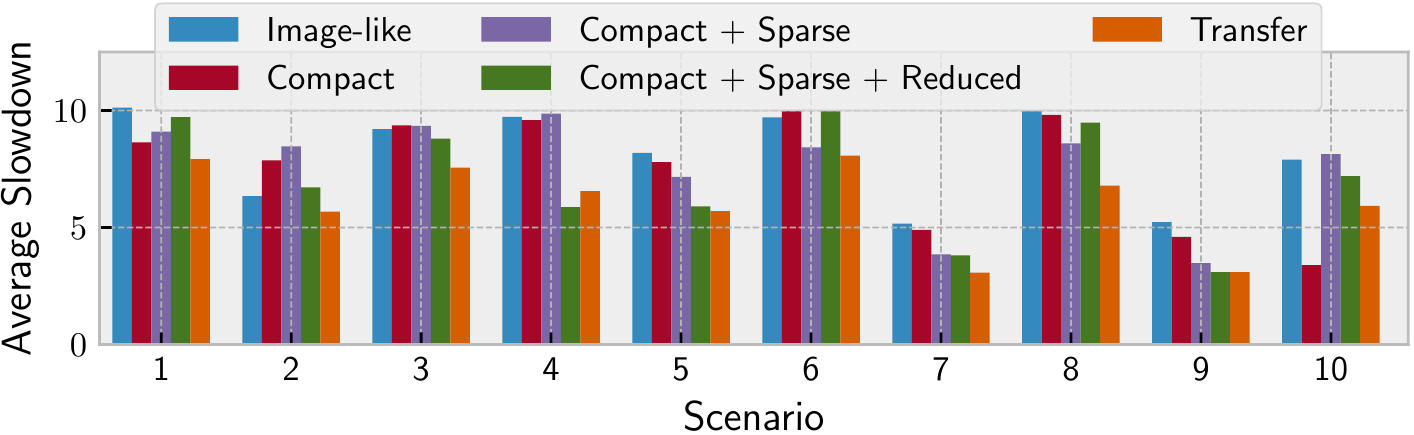}
  \caption{%
    Bar chart contrasting the performance of a transferred agent to agents
    trained specifically in their environments.
  }\label{fig:transfer}
\end{figure}

The compact \acp{MDP} proposed in this paper all have the characteristic of
having a state representation with a fixed size, which allows for
\emph{transfer} of learned weights between \acp{MDP}. Here, we consider
transfer the ability to change cluster configuration without the need for
retraining an agent from scratch, which is simply not possible when using the
image-like representation. In Figure~\ref{fig:transfer}, for example, we show
the performance of an agent trained in the bounded reward, event-based, compact
\ac{MDP} with 64 processors and with jobs of length 33 (the best agent in
Fig.~\ref{fig:average-slowdown-1}, corresponding to scenario 9) evaluated in a
compact environment \emph{without} event-based updates. With this same
agent, we were able to evaluate its performance in all different scenarios,
without the need for retraining. We see that, for the most part, slowdown is
kept low, and not only that: this agent outperformed other agents in 80\% of
scenarios (differences are statistically significant, with p-value $\le 1\%$,
except for scenario 9, since this is the same agent, and scenario 5, where the
test has low power to reject the null hypothesis). This highlights the
advantage of using a representation that allows for easy transfer between
agents, enabling good performance in a variety of cluster settings.

\newpage
\section{Conclusion}\label{sec:conclusion}

In this paper, we've filled a gap in the literature by analyzing the effects of
different \ac{MDP} design decisions on the behavior of \ac{RL} agents. In
particular, we experimented with resource management agents for job scheduling
in computing clusters, discussing cases in which a compact representation
outperforms a dense one, and vice-versa. We proposed a new state
representation, a transition function, and a reward function for an \ac{MDP}
studied in the literature, and we saw that these environments support
transferring agents between different cluster settings, while also keeping
agent memory consumption constant, and processing requirements stable.
We also saw that these compact representations are no worse than image-like
ones, and, thus, might be preferable when constant memory usage is
a requirement. Moreover, our results indicate that transferred agents may
outperform specialized agents in 80\% of the tested scenarios without the need
for retraining.

\bibliographystyle{plainnat}
\bibliography{references}

\begin{thebibliography}{16}
\providecommand{\natexlab}[1]{#1}
\providecommand{\url}[1]{\texttt{#1}}
\expandafter\ifx\csname urlstyle\endcsname\relax
  \providecommand{\doi}[1]{doi: #1}\else
  \providecommand{\doi}{doi: \begingroup \urlstyle{rm}\Url}\fi

\bibitem[Cunha et~al.(2017)Cunha, Rodrigues, Tizzei, and Netto]{cunha2017job}
Renato~L.F. Cunha, Eduardo~R. Rodrigues, Leonardo~P. Tizzei, and Marco~A.S.
  Netto.
\newblock {Job placement advisor based on turnaround predictions for HPC hybrid
  clouds}.
\newblock \emph{Future Generation Computer Systems}, 67:\penalty0 35 -- 46,
  2017.
\newblock ISSN 0167-739X.

\bibitem[de~Freitas~Cunha and Chaimowicz(2020)]{cunha2020towards}
Renato~Luiz de~Freitas~Cunha and Luiz Chaimowicz.
\newblock Towards a common environment for learning scheduling algorithms.
\newblock In \emph{2020 28th International Symposium on Modeling, Analysis, and
  Simulation of Computer and Telecommunication Systems (MASCOTS)}, pages 1--8,
  2020.

\bibitem[Domeniconi et~al.(2019)Domeniconi, Lee, and
  Morari]{domeniconi2019cush}
Giacomo Domeniconi, Eun~Kyung Lee, and Alessandro Morari.
\newblock {CuSH: Cognitive ScHeduler for Heterogeneous High Performance
  Computing System}.
\newblock In \emph{Proceedings of DRL4KDD 19: Workshop on Deep Reinforcement
  Learning for Knowledge Discovery (DRL4KDD)}, volume~12, 2019.

\bibitem[Fan et~al.(2021)Fan, Lan, Childers, Rich, Allcock, and
  Papka]{fan2021deep}
Yuping Fan, Zhiling Lan, Taylor Childers, Paul Rich, William Allcock, and
  Michael~E Papka.
\newblock Deep reinforcement agent for scheduling in hpc.
\newblock \emph{arXiv preprint arXiv:2102.06243}, 2021.

\bibitem[Feitelson and Rudolph(1996)]{feitelson1996toward}
Dror~G Feitelson and Larry Rudolph.
\newblock Toward convergence in job schedulers for parallel supercomputers.
\newblock In \emph{Workshop on Job Scheduling Strategies for Parallel
  Processing}, pages 1--26. Springer, 1996.

\bibitem[Liang et~al.(2016)Liang, Machado, Talvitie, and
  Bowling]{liang2016state}
Yitao Liang, Marlos~C. Machado, Erik Talvitie, and Michael Bowling.
\newblock {State of the Art Control of Atari Games Using Shallow Reinforcement
  Learning}.
\newblock In \emph{AAMAS}, 2016.

\bibitem[Mao et~al.(2016)Mao, Alizadeh, Menache, and Kandula]{mao2016resource}
Hongzi Mao, Mohammad Alizadeh, Ishai Menache, and Srikanth Kandula.
\newblock {Resource management with deep reinforcement learning}.
\newblock In \emph{Proceedings of the 15th ACM Workshop on Hot Topics in
  Networks}, pages 50--56, 2016.

\bibitem[Mnih et~al.(2016)Mnih, Badia, Mirza, Graves, Lillicrap, Harley,
  Silver, and Kavukcuoglu]{mnih2016asynchronous}
Volodymyr Mnih, Adria~Puigdomenech Badia, Mehdi Mirza, Alex Graves, Timothy
  Lillicrap, Tim Harley, David Silver, and Koray Kavukcuoglu.
\newblock {Asynchronous methods for deep reinforcement learning}.
\newblock In \emph{International conference on machine learning}, pages
  1928--1937, 2016.

\bibitem[Nishida et~al.(2018)Nishida, Costa, and Bianchi]{nishida2018control}
Cyntia Eico~Hayama Nishida, Anna Helena~Reali Costa, and Reinaldo Augusto
  da~Costa Bianchi.
\newblock Control of gene regulatory networks basin of attractions with batch
  reinforcement learning.
\newblock In \emph{2018 7th Brazilian Conference on Intelligent Systems
  (BRACIS)}, pages 127--132, 2018.

\bibitem[Raffin et~al.(2019)Raffin, Hill, Ernestus, Gleave, Kanervisto, and
  Dormann]{stable-baselines3}
Antonin Raffin, Ashley Hill, Maximilian Ernestus, Adam Gleave, Anssi
  Kanervisto, and Noah Dormann.
\newblock Stable baselines3.
\newblock \url{https://github.com/DLR-RM/stable-baselines3}, 2019.

\bibitem[Ramos et~al.(2020)Ramos, Silva, Araujo, Marcolino, and
  Nascimento]{ramos2020straight}
Washington Ramos, Michel Silva, Edson Araujo, Leandro~Soriano Marcolino, and
  Erickson Nascimento.
\newblock Straight to the point: Fast-forwarding videos via reinforcement
  learning using textual data.
\newblock In \emph{Proceedings of the IEEE/CVF Conference on Computer Vision
  and Pattern Recognition}, pages 10931--10940, 2020.

\bibitem[Schulman et~al.(2017)Schulman, Wolski, Dhariwal, Radford, and
  Klimov]{schulman2017proximal}
John Schulman, Filip Wolski, Prafulla Dhariwal, Alec Radford, and Oleg Klimov.
\newblock {Proximal policy optimization algorithms}.
\newblock \emph{arXiv preprint arXiv:1707.06347}, 2017.

\bibitem[Silver et~al.(2018)Silver, Hubert, Schrittwieser, Antonoglou, Lai,
  Guez, Lanctot, Sifre, Kumaran, Graepel, et~al.]{silver2018general}
David Silver, Thomas Hubert, Julian Schrittwieser, Ioannis Antonoglou, Matthew
  Lai, Arthur Guez, Marc Lanctot, Laurent Sifre, Dharshan Kumaran, Thore
  Graepel, et~al.
\newblock {A general reinforcement learning algorithm that masters chess,
  shogi, and Go through self-play}.
\newblock \emph{Science}, 362\penalty0 (6419):\penalty0 1140--1144, 2018.

\bibitem[Sutton et~al.(1999)Sutton, Precup, and Singh]{sutton1999between}
Richard~S. Sutton, Doina Precup, and Satinder Singh.
\newblock {Between MDPs and semi-MDPs: A framework for temporal abstraction in
  reinforcement learning}.
\newblock \emph{Artificial Intelligence}, 112\penalty0 (1):\penalty0 181--211,
  1999.
\newblock ISSN 0004-3702.
\newblock \doi{https://doi.org/10.1016/S0004-3702(99)00052-1}.

\bibitem[Tesauro(1994)]{tesauro1994td}
Gerald Tesauro.
\newblock {TD-Gammon, a self-teaching backgammon program, achieves master-level
  play}.
\newblock \emph{Neural computation}, 6\penalty0 (2):\penalty0 215--219, 1994.

\bibitem[Zhang et~al.(2020)Zhang, Dai, He, Bao, and Xie]{zhang2020rlscheduler}
Di~Zhang, Dong Dai, Youbiao He, Forrest~Sheng Bao, and Bing Xie.
\newblock Rlscheduler: an automated hpc batch job scheduler using reinforcement
  learning.
\newblock In \emph{SC20: International Conference for High Performance
  Computing, Networking, Storage and Analysis}, pages 1--15. IEEE, 2020.

\end{thebibliography}

\end{document}